\documentclass{article}

\usepackage{arxiv}

\usepackage[utf8]{inputenc} 
\usepackage[T1]{fontenc}    
\usepackage{hyperref}       
\usepackage{url}            
\usepackage{booktabs}       
\usepackage{amsfonts}       
\usepackage{nicefrac}       
\usepackage{microtype}      
\usepackage{lipsum}
\usepackage{graphicx}
\usepackage{soul}
\usepackage{bbold}
\soulregister\cite7
\soulregister\ref7
\soulregister\pageref7

\usepackage{amsmath}

\usepackage{graphicx}
\usepackage{threeparttable}
\usepackage{amsmath}
\usepackage{algorithm,algorithmic}
\usepackage{multirow}
\usepackage{comment}
\usepackage{color,soul}
\usepackage{empheq}
\usepackage[theorems,skins]{tcolorbox}
\usepackage{amssymb,amsfonts}
\usepackage{textcomp}
\usepackage{subcaption}
\usepackage{threeparttable}

\newcommand{\tensor}{\mathsf}
\newcommand{\vct}{\mathbf}

\title{EmergencyNet: Efficient Aerial Image Classification for Drone-Based Emergency Monitoring using Atrous Convolutional Feature Fusion}

\author{
  Christos Kyrkou\thanks{ckyrkou@gmail.com, www.christoskyrkou.com
  } \\
  KIOS Research and Innovation Center of Excellence\\
  University of Cyprus\\
  1 Panepistimiou Avenue, Nicosia Cyprus \\
  \texttt{kyrkou.christos@ucy.ac.cy} \\
   \And
 Theocharis Theocharides \\
  Department of Electrical and Computer Engineering\\
   KIOS Research and Innovation Center of Excellence\\
  University of Cyprus\\
  1 Panepistimiou Avenue, Nicosia Cyprus \\
  \texttt{ttheocharides@ucy.ac.cy} 
}

\begin{document}
\maketitle

\begin{abstract}
Deep Learning based algorithms can provide state-of-the-art accuracy for remote sensing technologies such as Unmanned Aerial Vehicles (UAVs)/drones, potentially enhancing their remote sensing capabilities for many emergency response and disaster management applications. In particular, UAVs equipped with camera sensors can operating in remote and difficult to access disaster-stricken areas, analyze the image and alert in the presence of various calamities such as collapsed buildings, flood, or fire in order to faster mitigate their effects on the environment and on human population. However, the integration of deep learning introduces heavy computational requirements, preventing the deployment of such deep neural networks in many scenarios that impose low-latency constraints on inference, in order to make mission-critical decisions in real-time. To this end, this paper focuses on the efficient aerial image classification from on-board a UAV for emergency response/monitoring applications. Specifically, a dedicated Aerial Image Database for Emergency Response (AIDER) applications is introduced and a comparative analysis of existing approaches is performed. Through this analysis a lightweight convolutional neural network (CNN) architecture is proposed, referred to as \textit{EmergencyNet}, based on atrous convolutions to process multiresolution features and capable of running efficiently on low-power embedded platforms achieving upto $20\times$ higher performance compared to existing models with minimal memory requirements with less than $1\%$ accuracy drop compared to state-of-the-art models.
\end{abstract}


\section{Introduction}\label{sec:intro}
Over the past few years Unmanned Aerial Vehicles (UAVs)/drones have gained considerable interest as a remote sensing platform for various practical applications, such as traffic monitoring \cite{Kyrkou:ICCE:2018}, search and rescue \cite{PetridesICUAS2017}, and precision agriculture \cite{PrAgri:Murugan:Garg}, and satelite imagery processing \cite{Zhao:JSTARS:2017}. Recent technological advances such as the integration of camera sensors provide the opportunity for new UAV applications such as detect, monitor and analyze passive and active threats and hazards at incident scenes (e.g., fire spots in forested areas, flooding threat, road collisions, landslide prone areas) \cite{CV_UAV_Survey:Abdulla:2018}. In addition, due to their small size UAVs offer fast deployment and can thus be in-the-loop of mission critical decisions to better manage the available resources and improve risk assessment, prevention, and mitigation \cite{PetridesSmartCities2017}. However, there is a unique set of constraints that need to be addressed due to the fact that a UAV has to operate in disaster-stricken areas where remote communication to cloud services may not be possible to be established and high-end infrastructure is not available. As a result, a higher level of autonomy is required to ensure operational efficiency and real-time analysis. In such cases an autonomous UAV relies heavily on its on-board sensors and microprocessors to carry out a given task without requiring the feed to be send to a central ground station. Furthermore, the autonomous operation of UAVs by combining path planning algorithms with automated on-board visual processing enables them to cover a larger area in less time \cite{PetridesICUAS2017}. However, on-board  processing comes with its own set of challenges due to the  limited computational resources and low-power constraints which are necessitated by the low-payload capabilities of UAVs. As such, the computational efficiency of the underlying computer vision algorithm plays a key role in enabling autonomous UAVs to detect hazards at incident scenes in real-time. 

Deep learning algorithms such as Convolutional Neural Networks (CNNs) have been widely recognized as a prominent approach for many computer vision applications (image/video recognition, detection, and classification) and have shown remarkable results in many applications \cite{SateliteImageClassification:Maggiori:2016,Hohman:2018:VisualAnalyticsDeepLearning:TVCG,Cheng:WhenDLmeetsML:2018:TGRS}. Hence, there are many benefits stemming from using deep learning techniques in emergency response and disaster management applications to retrieve critical information in a timely-fashion and enable better preparation and reaction during time-critical situations and support in-the-loop decision-making processes \cite{Nguyen2016ApplicationsOO}. Prior works have demonstrated how deep learning approaches can overcome traditional machine learning methods with hand-crafted features through the use of transfer learning where a pretrained convolutional neural network is used as a feature extractor and one or more layers are added on top to perform the classification for the new task \cite{Razavian:CNNtransfer:CVPRW:2014}. Even though CNNs are increasingly successful at classification tasks, their inference speed on embedded low-power platforms such as those found on-board UAVs is hindered by the high computational cost that they incur  \cite{DATELowPowerImage2018,Shafique.RoadMap.DATE2018}, especially when considering the need to run multiple vision tasks on the same platform. As such, for many applications local embedded processing near the sensor is preferred over the cloud due to privacy or latency concerns, or operation in remote areas where there is limited or even no connectivity. In addition, purpose built small networks can provide the necessary accuracy and performance for niche applications where abundant data is not available and computational constraints are imposed. Also, beyond the computational efficiency they are faster to go through training iterations and more easily updatable over-the-air. Hence, using a small CNN that is amicable for near-sensor (edge) processing to perform the aerial scene classification on board a UAV becomes a very attractive and sensible alternative to standard approaches. 

\begin{figure}[t]
	\centering
	\includegraphics[width=0.7\columnwidth]{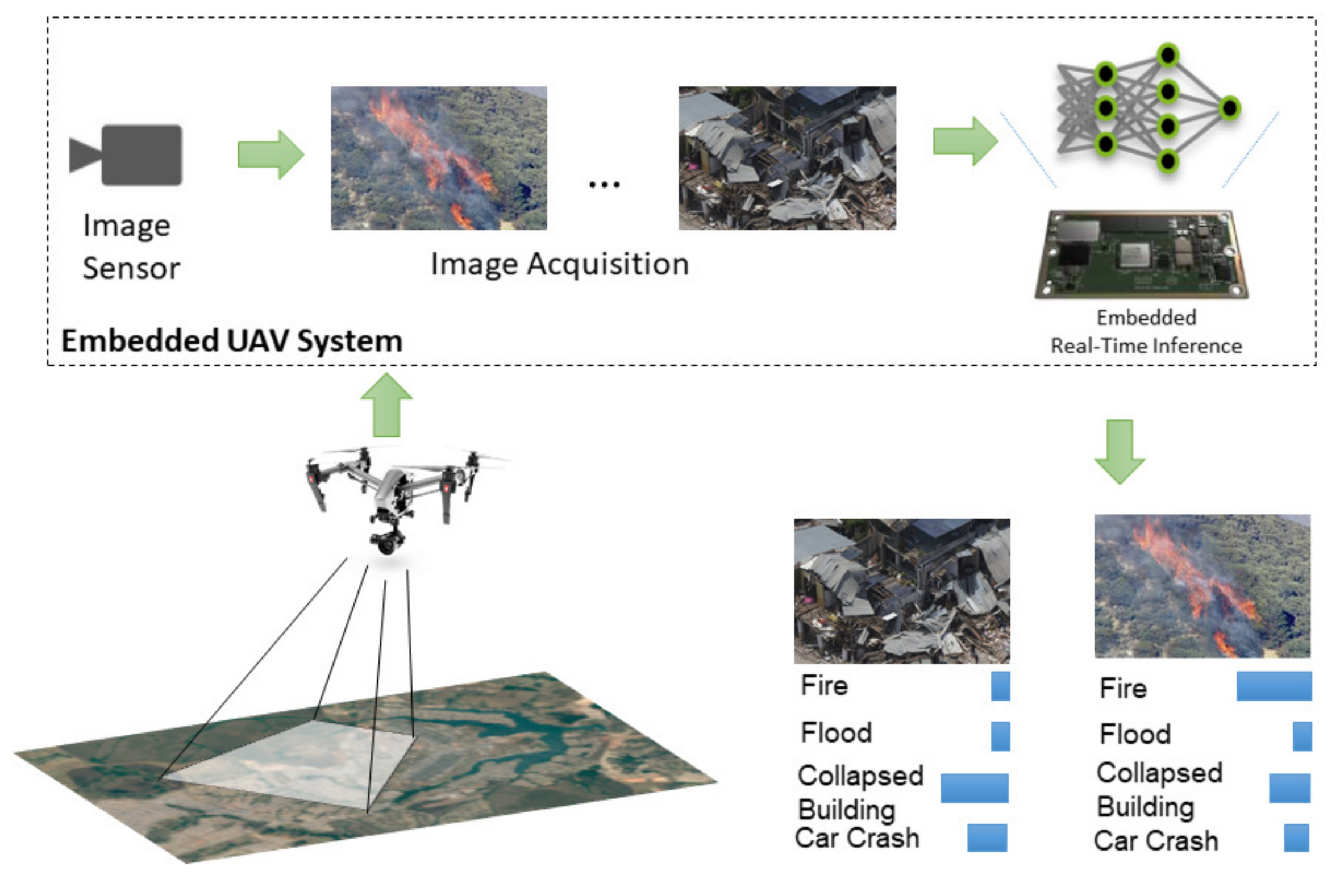}
	\caption{Application scenario for use of UAVs that utilize algorithms based on deep learning models that analyse video footage in real-time to characterizing the current situation and alert in the presence of any danger.}
	\label{fig:scenario}
\end{figure}

This work addresses the problem of on-board aerial scene classification which is to automatically assign a semantic label to characterize the aerial image that the UAV captures \cite{Wang:ASC:TGRS:2017}. With respect to emergency response applications these labels correspond to a danger or hazard that has occurred. Such a system can be deployed into UAVs for automated monitoring and inspection to enhance preparedness and provide rapid situational awareness. The specific use-case under consideration is that a UAV will follow a predetermined path as shown in \cite{PetridesICUAS2017}) and will continuously analyze the frames it receives from the camera through its embedded platform and alert for any potential hazards or dangerous situations that it recognizes as shown in Fig. \ref{fig:scenario}. The objective of this work is to enhance the real-time perception capabilities in such scenarios through the development of a CNN model that provides the best trade-off between accuracy and performance and can operate on embedded hardware that is on-board the UAV or its mobile control station. The preliminary work in \cite{Kyrkou_2019_CVPR_Workshops}  is significantly extended by proposing a lightweight CNN, referred to as \textit{EmergencyNet}, based on depthwise atrous convolutions enabling it to gather multi-resolution features in a computationally efficient manner thus managing to provide adequate trade-off between accuracy and speed for the problem of emergency response monitoring with UAVs. Further, the dataset is significantly enlarged and perform an extended comparison with other models and also propose further optimizations to improve performance in certain use-cases.

The main contributions of this work are summarized as follows:
\begin{itemize}
	\item A dedicated database for the application of aerial image\footnote{The dataset will be released at https://github.com/ckyrkou/AIDER/} classification for emergency response which contains a larger number of images compared to existing datasets.
	\item A novel and computationally efficient CNN (referred to as \textit{EmergencyNet}) is proposed that combines multi-resolution depthwise convolutions to simultaneously provide near state-of-the-art accuracy ($\sim95.7\%$) while being up to $\sim20\times$ faster,and suitable for low-cost low-power devices.
	\item Study and analyze the impact of different CNN architectures for the task of aerial scene classification of disasters and evaluation in terms of accuracy, inference speed, and memory.
	\item Evaluation of the different models using an actual experimental setup consisting of a UAV with two processing options an embedded computing platform and a mobile ground station. 
\end{itemize}

The rest of this paper is structured as follows. Section \ref{sec:background} provides the background on convolutional neural network architectures and outlines previous work on the area of aerial image classification for emergency response and disaster management. Section \ref{sec:approach} provides details on the constructed dataset for aerial image classification of disasters as well as the techniques used to develop the proposed network. Section \ref{sec:results} presents an analysis of the different models and techniques as well as evaluation on a real experimental UAV platform.  Finally, Section \ref{sec:conc} provides concluding remarks and discusses directions for future work in this area.

\section{Background and Related Work}\label{sec:background}

\subsection{Convolutional Neural Network Architectures and design Approaches}\label{subsec:CNNs}

Convolutional neural networks (CNNs) are a biologically inspired machine learning algorithm that can be trained to perform a variety of image detection, recognition and segmentation tasks. CNNs are composed of multiple layers that are trained using stochastic gradient descent with back-propagation in order to learn hierarchical represenations of visual data. In the last decade, a lot of progress has been made on CNN-based classification systems. Numerous architectures have been proposed by the deep learning community fuelled by the need to perform even better in image classification tasks such as the ImageNet Large Scale Visual Recognition Competition (ILSVRC). Typically, the structure of a CNN comprises a feature extractor stage followed by a classification layer. Some of the most important architectures are highlighted next. The reader is referred to the individual papers for more details.

\textbf{VGG16 \cite{VGG.2014}:} The VGG network has become a popular choice when extracting CNN features from images. This particular network contains $16$ CONV/FC layers and appealingly, is characterized by its simplicity. It is comprised only of $3\times 3$ convolutional layers stacked on top of each other with an increasing depth of $2$ with pooling layers in between to reduce the feature map size by a factor of $2$; and with $2$ fully-connected layers at the end, each with $4,096$ neurons. A final dense layer is equal to the number of classes is used for the final classification. A downside of the VGGNet is that it is more expensive to evaluate, and uses a lot of parameters and consequently memory ($\sim140$MB). 

\textbf{ResNet \cite{ResNet.2015}:} This network introduced the idea of residual learning in order to train even deeper CNNs, where the input to a convolution layer is propagated and added to the output of that layer after the operation, thus the network effectively learns residuals. However, it's gain in accuracy comes at a cost of both memory demands as well as execution time. ($\sim102$MB)

\textbf{Inception \cite{SzegedyInception2014} \& Xception \cite{Xception.2016}}:The main contribution of this architecture is that it combines many different convolution filters (e.g., $1\times1, 3\times3, and 5\times5$) into a multi-level feature extractor. The output of these filters at the same network level are stacked along the channel dimension  before being fed into the next layer. This module is referred to as \textit{inception} module in the network. The specific architecture referred to as \textit{Inception V3} comes form the work of Szegedy et al. \cite{Szegedy:CVPR:2015:IncV3} which proposed updates to the original inception module to further boost accuracy. The weights for Inception V3 are smaller than both VGG and ResNet, coming in at $\sim96$MB but can still be considered too large for embedded applications. For this reason a more optimized variant of the inception family was proposed called \textit{Xception} where the idea of separable convolutions was proposed in an attempt to decrease the computational complexity. A convolution is separated to a depth-wise separable convolution where a filter is applied independently on each channel and then these results are combined through point-wise convolutions. Herein, the focus will be on the \textit{Xception} architecture due to its comparatively higher computational efficiency \cite{Xception.2016}.  

\textbf{MobileNet \cite{MobileNets.2017,MobileNetsV2.2018,Howard_2019_ICCV}:} Utilizing the idea of separable convolutions the MobileNet family of models manage to offer state-of-the-art performance with reduced computational cost. Essentially, the architecture is a streamlined version of the Xception network that applies a single filter at each input in contrast to the more complicated inception module. Thus is designed can be easily parametrized and optimized for mobile applications. In the latest iteration of MobileNets \cite{Howard_2019_ICCV} the authors utilize neural search techniques to further optimize the model architecture, especially for semantic segmentation tasks.

\textbf{SqueezeNet \cite{SqueezeNet:2016}:} This is a CNN designed to be parameter efficient. It utilizes a fire module that combines the output of different convolutional layers. It also fetures a bottleneck layer that reduces the dimensionality of the input feature map to subsequently reduce the computational cost.

\textbf{ShuffleNet \cite{ShuffleNetV2:2018}:} This approach relies on utilizing different groups of filters each working on a separate channel group of the input feature map which reduces the parameter count of the network. In order to then relate the different channels together it utilizes a process called channel shuffling.

\textbf{EfficientNet \cite{EfficientNet:2019}:} A new CNN scaling approach is proposed in this work that uniformly scales all dimensions of depth/width/resolution in a principled manner in order to achieve better accuracy and efficiency. The approach can be utilized with any backbone architecture such as \textit{MobileNets} and \textit{ResNets}. In addition, autoML framework is used to automatically search for an optimized architecture. Overall, efficientNets have found to surpass the accuracy of larger models such as \textit{VGG} with an order of magnitude fewer parameters and FLOPS. However, the evaluation of EfficientNets was limited to cloud-centric processing systems.

\subsection{Related work on image classification for emergency response and disaster management}\label{subsec:rw}

In this section some relevant works for the problem of aerial image classification for emergency response and disaster management are described, some of which also target remote sensing with UAVs. Different methods have been proposed over the years to detect various disasters in images such as image-processing-based with thresholds to perform pixel-level classification \cite{YuanICUASFire2015}, Gaussian mixture models which require empirical tuning \cite{Toreyin:2006:CVB:1131923.1131930}, and Support Vector Machines which are cosnidered slow for real-time applications \cite{ChulSVMFire2009}. The success of deep learning and CNNs in particular for different kinds of image analysis tasks has also led the research community to investigate their suitability for such applications. Such approaches have first been proposed for ground robots such as the work in \cite{Li2017DistributedDisManagement} and later also used to interpret aerial images \cite{Mnih2012LLANoisy}. 

Deep learning has gained a prominent role as an approach for aerial image classification for emergency response and disaster management applications due to its higher classification accuracy and generalization capabilities. In \cite{Kim.Fire.2016} the authors propose a cloud based deep learning approach for fire detection with UAVs. The detection using a custom convolutional neural network (similar in strcuture to VGG16) which is trained to discriminate between fire and non-fire images of $128\times128$ resolution. The system works by transmitting the video footage from a UAV to a workstation with an NVIDIA Titan Xp GPU where the algorithm is executed. Of course, in scenarios with limited connectivity there would be difficulties in applying this approach. Overall, the proposed approach achieves an accuracy in the range of $81-88\%$ for this task.

In \cite{Bejiga.Avalanche.2017} a method is proposed for detecting objects of interest in avalanche debris using the pretrained inception Network for feature extraction and a linear Support Vector Machine for the classification. They also propose an image segmentation method as a preprocessing technique that is based on the fact that the object of interest is of a different color than the background in order to separate the image into regions using a sliding window. In addition, they apply post-processing to improve the decision of a classifier based on hidden Markov models. The application is executed on a desktop computer and not on an embedded device, with clock speed of 3GHz and 8GB RAM average a performance of 5.4 frames per second for $224\times 224$ images. The accuracy was between $72-97\%$.

Similarly, the work in \cite{Sharma.Fire.2017} also targets fire detection application with deep learning. Specifically, two pretrained convolutional neural networks are used and compared, namely VGG16 \cite{VGG.2014} and Resnet50 \cite{ResNet.2015} as base architectures to train fire detection systems. The architectures are adapted by adding fully connected layers after the feature extraction to measure the classification accuracy. The different models average an accuracy of $\sim91\%$ for a custom database with an average processing time of $1.35$ seconds on an NVIDIA GeForce GTX $820$ GPU.

The work in \cite{Zhao.SaliencyFire.2018}  proposes an approach for wildfire detection from UAV platform. The overall approach comprises of a convolutional neural network called \textit{Fire\_Net} consisting of $15$ layers with an architecture similar to the \textit{VGG16} network with $8$ convolutional, $4$ max-pooling, and $2$ fully connected layers for recognizing fire in $128\times 128$ resolution images. It is accompanied by a region proposal algorithm that extracts image regions from larger resolution images so that they can be classified by the neural network. The training of the system was performed on an NVIDIA GeForce $840$M GPU, while the overall accuracy is $\sim98\%$ and the average performance is $24$ frames-per-second on the particular GPU platform for $128\times 128$ resolution images and without considering the overhead of the region proposal and region selection algorithms.  

Perhaps the most relevant related work in terms of application domain is that of Kamilaris et al \cite{Kamilaris.DisManUAV.2017} where a deep convolutional neural network is trained to classify aerial photos in one of $5$ classes corresponding to natural disasters. The VGG \cite{VGG.2014} network is used as the base feature extraction and a fully connected is placed on top of it to perform the transfer learning for the new task. An accuracy of $91\%$ is achieved for a custom test set and on average less than $3$ seconds are needed to process an image of $224\times224$ on an Intel Core i7 machine. Even though the application is similar, in this work the UAV on-board system is considered as the processing platform which has additional constraints with the objective of increasing its autonomy and real-time processing capabilities.

From the literature analysis it is clear that existing approaches targeting aerial scene classification for emergency response and disaster management with UAVs have shown really good results using deep CNN architectures as the main classification approach. In their majority they use existing pretrained networks which adapt through transfer learning for the classification of a single event and primarily utilize desktop-class systems as the main computational platform that remotely process the UAV footage on GPUs. However, in certain scenarios the communication latency and connectivity issues may hinder the performance of such systems necessitating higher autonomy levels for the UAV and on-board processing capabilities \cite{Sze.MLchallOppo.2016}. Moreover, the computing limitations of embedded platforms constitute the use of existing algorithms targeting desktop-class systems infeasible. To that end the contribution of this paper with regards to state-of-the-art is the evaluation of existing approaches for the task of aerial scene classification for emergency response and disaster management and the introduction of an efficient convolutional neural network suitable for embedded platforms such as UAVs that is capable for on-board classification of multiple disaster events.

\section{Deep Learning for Aerial Disaster-Event Classification}\label{sec:approach}

This section outlines the process of developing an efficient convolutional neural network suitable for embedded platforms for classifying aerial images from a UAV for emergency response and disaster management applications. Specifically, it details how the training set for this problem was collected, and the different networks used for analysis and comparison both through transfer learning of pretrained networks and custom networks along with the design choices made to develop them. Finally, the details of the training process are provided.

\subsection{Dataset Collection}\label{subsec:aider}

Training a CNN for aerial image classification for emergency response and disaster management applications first requires collecting a suitable dataset for this task. To the best of our knowledge there is no widely used and publicly available dataset for emergency response applications. As such, a dedicated database for this task is constructed referred to as \textit{AIDER} (\textbf{A}erial \textbf{I}mage \textbf{D}ataset for \textbf{E}mergency \textbf{R}esponse Applications). The dataset construction involved manually collecting all images for four disaster events, namely  \textit{Fire/Smoke}, \textit{Flood}, \textit{Collapsed Building/Rubble}, and \textit{Traffic Accidents}, as well as one class for the \textit{Normal} case.  Visually similar images such as for example active flames and smoke are grouped together. A finer-level classification is possible but is left as future work. 

The aerial images for the disaster events were collected through various online sources (e.g. google images, bing images, youtube, news agencies web sites, etc.) using the keywords "Aerial View" or "UAV" or"Drone" and an event such as "Fire","Earthquake","Highway accident", etc. 
Images are initially of different sizes but are standardized prior to training. All images where manually inspected to first contain the event that was of interested and then to have the event centered at the image so that any geometric transformations during augmentation would not remove it from the image view. During the data collection process the various disaster events were captured with different resolutions and under various condition with regards to illumination and viewpoint.  Finally, to replicate real world scenarios the dataset is imbalanced in the sense that it contains more images from the \textit{Normal} class. Of course, this can make the training more challenging, however, an appropriate strategy is followed to combat this during training which will be detailed in the following sections. 

\begin{figure}[t]
	\centering
	\includegraphics[width=0.7\linewidth]{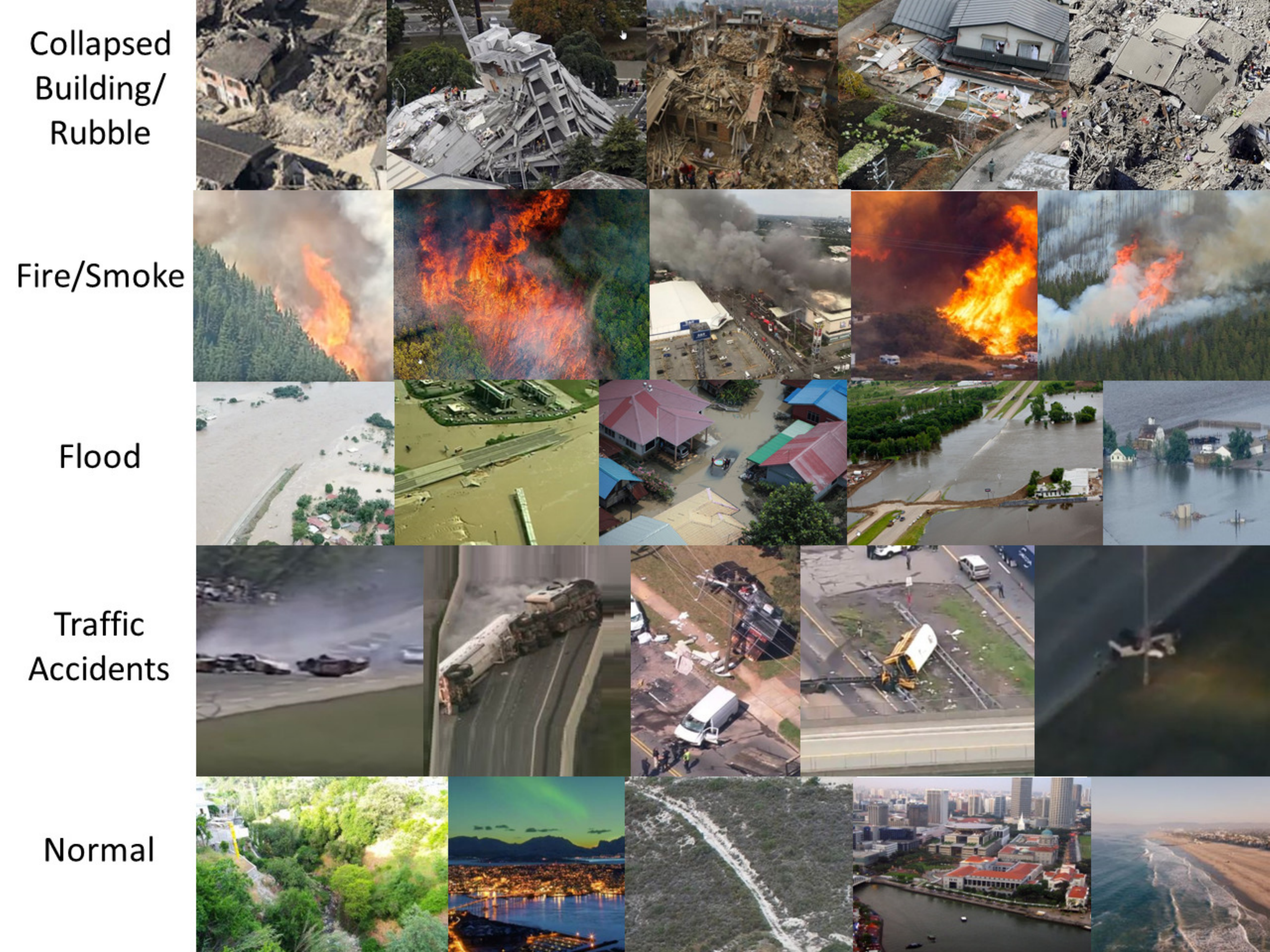}
	\caption{\textbf{A}erial \textbf{I}mage \textbf{D}ataset for \textbf{E}mergency \textbf{R}esponse (AIDER) Applications: Example images from the database by class.}
	\label{fig:dataset}
\end{figure}

The operational conditions of the UAV may vary depending on the environment, as such it is important that the dataset does not contain only "clean" and "clear" images. In addition, data-collection can be time-consuming and expensive. Hence to further enhance the dataset a number random augmentations are probabilistically applied to each image prior to adding it to the batch for training. Specifically these are geometric transformations such as rotations, translations, horizontal axis mirroring, cropping and zooming, as well as image manipulations such as illumination changes, color shifting, blurring, sharpening, and shadowing. In addition, sample pairing is also employed to mix images together and further enhance the training set \cite{samplePairing:2018}. Each transformation is applied with a random probability which is set in such as way to ensure that not all images in a training batch are transformed so that the network does not capture the augmentation properties as a characteristic of the dataset. The objective of all this transformations is to combat overfitting and increase the variability in the training size to achieve a higher generalization capability. Some samples from the dataset can be seen in Fig. \ref{fig:dataset}. Overall, with respect to the related works that consider multiclass problems (e.g., \cite{Kamilaris.DisManUAV.2017}) almost $17\times$ more data were collected. The dataset does not contain the amount of images found in common benchmarks such as ImageNet and CIFAR however, it is much more challenging to encounter such real-life events and capture adequate data. As such, augmentations techniques were utilized to enhance the initial dataset even further. This, in our opinion, appoints \textit{AIDER} to a valuable complementary data source for developing and benchmarking data-driven methodologies for emergency response and disaster monitoring applications.

%
%
%
%
%
%
%
%

\begin{table}[]
	\centering
	\caption{Summary of the \textbf{A}erial \textbf{I}mage \textbf{D}ataset for \textbf{E}mergency \textbf{R}esponse Applications (AIDER)}
	\label{table:dataset}
	\resizebox{\textwidth}{!}{%
		\begin{tabular}{l|c|c|c|c}
			\textbf{Class}                     & \textbf{Train Set} & \textbf{Validation Set} & \textbf{Testing Set} & \textbf{Total Per Class} \\ \hline
			\textit{Collapsed Building/Rubble} & 400                & 100                     & 200                  & 700                      \\ \hline
			\textit{Fire/Smoke}                & 420                & 110                     & 220                  & 740                      \\\hline
			\textit{Flood}                     & 400                & 100                     & 200                  & 700                      \\\hline
			\textit{Traffic Accidents}         & 400                & 100                     & 200                  & 700                      \\\hline
			\textit{Normal}                    & 2700               & 1000                    & 2000                 & 5700                     \\\hline
			\textbf{Total per Set}             & 4320               & 1410                    & 2810                 & \textbf{Overall: 8540}  
		\end{tabular}%
	}
\end{table}

\subsection{CNNs for Aerial Disaster Classification}\label{subsec:cnn_models}

To identify the best structure of the CNN that will perform the aerial image classification a number of different networks was developed using two different approaches. The overall objective of this process is to explore the performance-accuracy trade-offs between these networks. First, transfer learning is employed to train the networks outlined in Section \ref{subsec:CNNs} which correspond to the methodology used in prior works. Furthermore, new network structures are designed and trained from scratch specifically for this task. The reasoning behind the latter approach is that it allows making those design choices that lead to more efficient networks that are fast to execute on embedded platforms and at the same time maintain the accuracy of larger networks.

\subsubsection{Transfer Learning Networks}\label{subsubsec:pretrained}
For transfer learning different networks from the literature, \textit{VGG16}\cite{VGG.2014}, \textit{ResNet50}\cite{ResNet.2015} \textit{FireNet}\cite{Zhao.SaliencyFire.2018}, \textit{MobileNets}\cite{MobileNets.2017,MobileNetsV2.2018,Howard_2019_ICCV}, \textit{Xception}\cite{Xception.2016}, \textit{ShuffleNet}\cite{ShuffleNetV2:2018}, \textit{EfficientNet} \cite{EfficientNet:2019}, and \textit{SqueezeNet}\cite{SqueezeNet:2016}. The majority of these networks hace also been used in related works \cite{Kamilaris.DisManUAV.2017,Sharma.Fire.2017,Kim.Fire.2016}. 

The feature extraction part is frozen for each of these networks, applying all necessary preprocessing steps to the input image, and add a classification layer on top similar to prior works. In contrast to other works a global (per feature-map) average-pooling layer is applied prior to the dense layers followed by a softmax classification layer at the end. The average pooling reduces the parameter count and the subsequent computational and memory requirements and has shown to perform equally as well with the traditional approaches \cite{Lin:2013:NiN:GAP}. Hence, the pretrained models used for comparison are inherently more efficient in terms of memory and operations that the networks used in the literature for this task, which utilize fully connected layers.

\subsubsection{Custom Networks}\label{subsubsec:custom}

Fine-tuning pretrained networks has some critical limitations. The preterained networks have different degrees of sensitivity depending on how the dataset is similar to the large-scale dataset used (e.g., ImageNet) \cite{Simon:CVPR:2019ImageNet:Features}. The larger and deeper networks attained through transfer learning may not be suited for resource-constraint systems such as UAV platforms, which impose limitations of the size of the platform, the weight, and its power envelope. Furthermore, it is inconvenient to change the architecture of existing networks since the pretraining should be re-conducted on the large-scale dataset (e.g., ImageNet), requiring high computational cost. For this reason there is a need to design specialized networks that are inherently computationally efficient to eliminate the aforementioned limitations \cite{Zhu:CVPR2019:ScratchDet}. 

The design space is explored by focusing on the layer configurations, type and connectivity. Consequently different networks are developed to better understand the trade-offs involved in the design choices. There are some systematic design choices that are made across the different network configurations. As a primary building block an architecture is developed that relies on fusing features produced by atrous convolutions of different degrees of dilation \cite{AtrousConv:2018:ECCV} and such architectures have primarily been used for segmentation tasks. They are employed herein as a means to simultaneously learn features at various resolutions more efficiently thus facilitating UAV applications which encounter areas of interest at different scales. The proposed architecture allows for flexible aggregation of the multi-scale contextual information while keeping the same resolution and reduced number of parameters. The Atrous Convolutional Feature Fusion (ACFF) block is detailed next.

\begin{figure*}[t]
	\centering
	\includegraphics[width=0.95\linewidth]{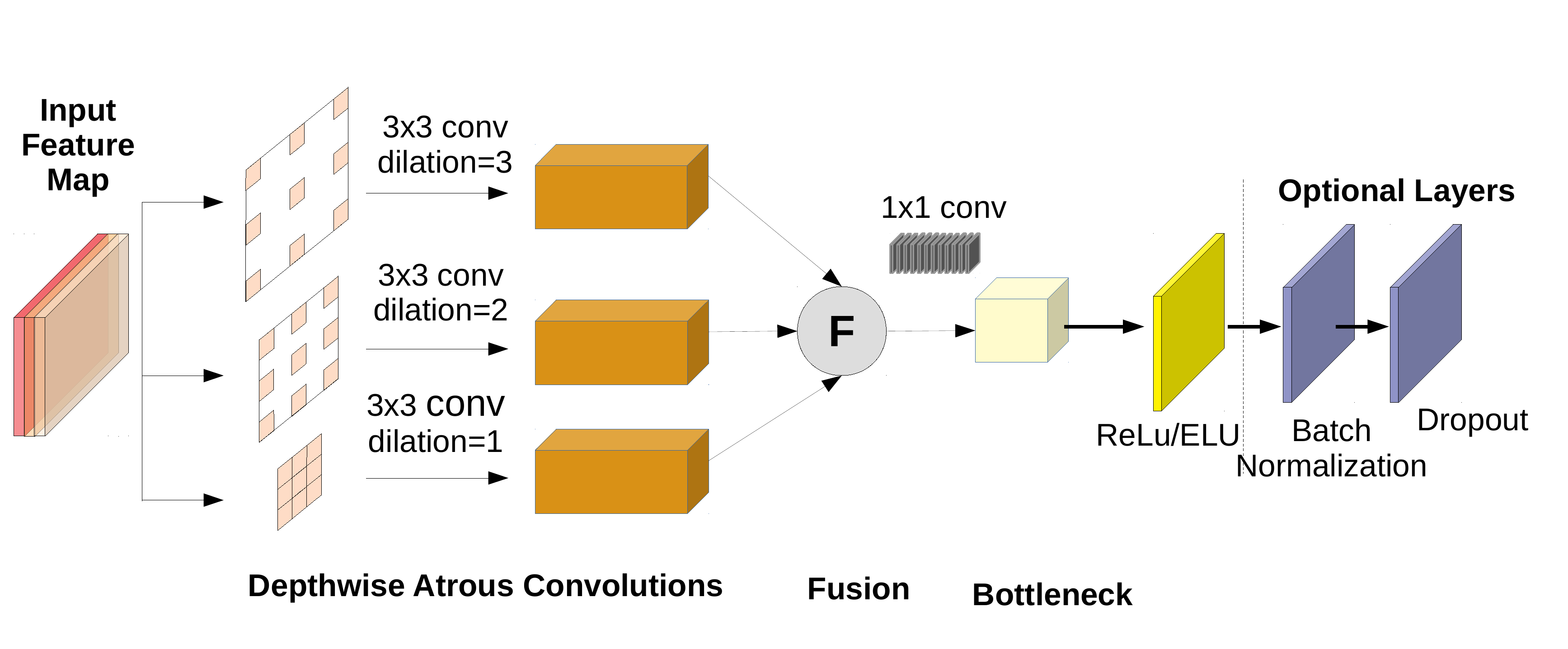}
	\caption{Basic components of multi-resolution Atrous Fusion Block.}
	\label{fig:ACFF}
\end{figure*}

\begin{figure}[t]
	\centering
	\includegraphics[width=0.7\linewidth]{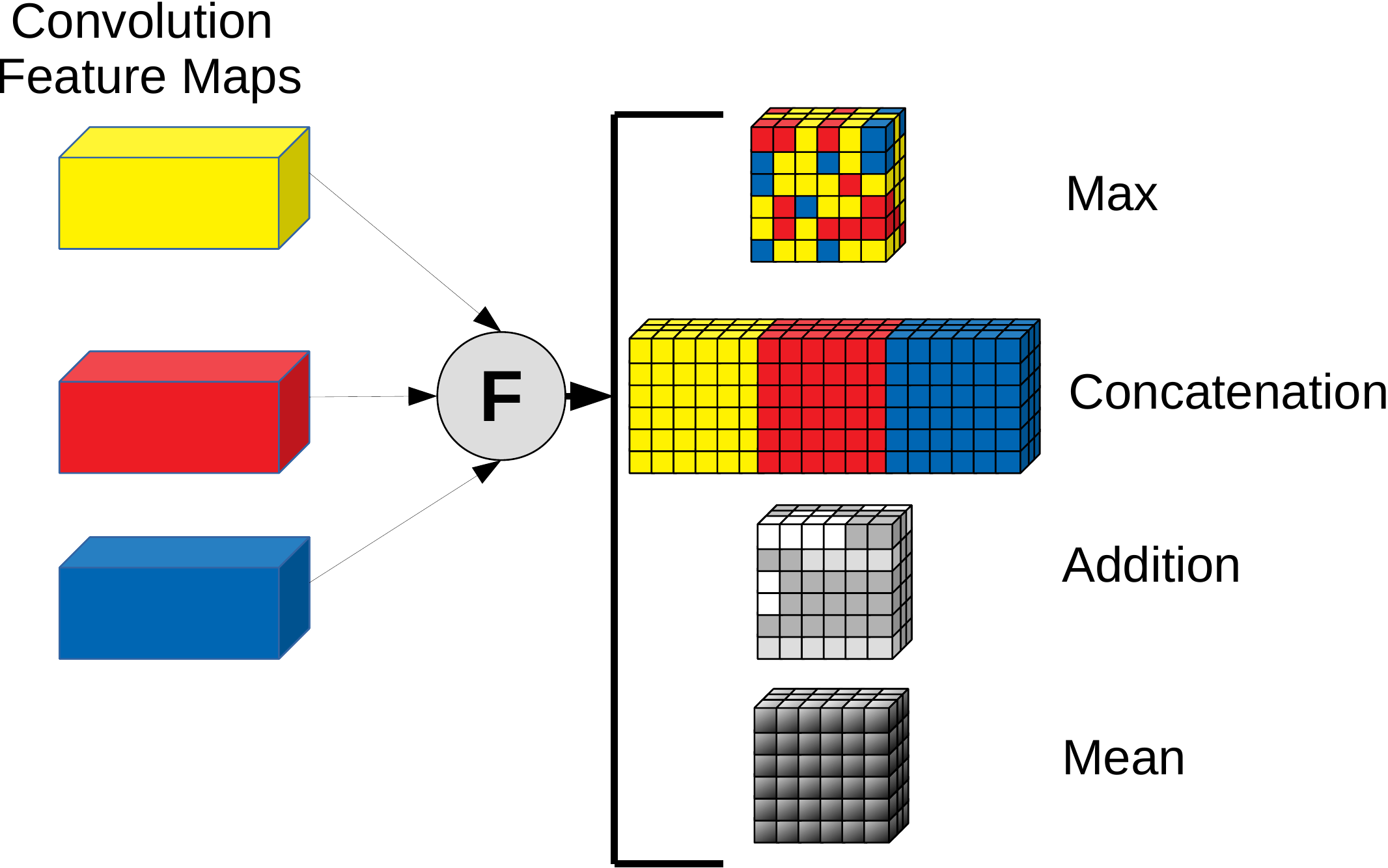}
	\caption{The different fusion schemes tested for the network configurations. The fusion nodes are instantiated after the atrous convolution blocks with element-wise add,mean,max operations, and a concatenation operation.}
	\label{fig:fusion}
\end{figure}

\subsection{Atrous Convolutional Feature Fusion (ACFF)}\label{acff}
Atrous (also called dilated) convolutions \cite{AtrousConv:2018:ECCV} can capture and transform images at different resolutions depending on the \textit{dilation rate} which determines the spacing between the kernel points, effectively increasing their receptive field without increasing the parameter count. Hence, it can be used to incorporate larger context to the model. The proposed block (Fig. \ref{fig:ACFF}) computes multiple such atrous convolutional features ($\tensor{U_d}$) for the same input map as shown in Eq. (\ref{eq:weight}) across different dilation rates $d$. Each atrous convolution is factored into depth-wise convolution that performs light-weight filtering by applying a single convolutional kernel per input channel to reduce the computational complexity. Then some form of fusion takes place to merge the different features together as shown in Eq. \ref{eq:fusion}. The intuition is to take advantage of the different dilation rates since one path may peek up features that another may have missed due to changes in object/region resolution. It is important to note that the weights are not shared between paths and each learns different weights $\vct{w_d}$ that may be more useful. Another advantage of using atrous convolutions stems from the fact that the same number of parameters and computations are needed regardless of the resolution. Each atrous convolution acts on the same feature map but at a different spatial resolution; starting from a dilation rate of $1$ and filter size of $3\times3$ (i.e., no spacing) and going up to $3$ which is equivalent to $7\times7$ receptive field effectively looking at the same area with different kernels but with less number of parameters utilized as shown in Table \ref{tab:dilatedconv}. 

An essential part of optimized CNNs is reducing not only the spatial size of feature maps but also the channel dimensions. Hence, prior to the atrous convolutions the input feature map channels are halved. This makes it possible to have multiple branches for atrous convolution without significantly impacting the performance. The depth reduction factor is a hyperparameter that can be further tuned depending on the requirements. Therefore, a bottleneck layer is utilized to reduce the number of channels after reshaping without changing the spatial size of the feature maps.

The atrous convolutional features at different dilation rates need to be combined together to allow the unit to learn from representations from a large effective receptive field. Four different fusion maps are examined \textit{maximum}, \textit{addition}, \textit{concatenation}, and \textit{averaging} as shown in Fig. \ref{fig:fusion}. The fusion mechanism is then followed by $1\times1$ convolutions and activation that non-linearly combine channel features together and projects them into a higher dimensional space. Finally, each atrous convolutional block is followed by batchnormalization and an activation. 

\begin{align}
	& \tensor{U_d}=\sum_{i=1}^{M}{\sum_{j=1}^{N}{\tensor{X}(m+d\times i,n+d\times j)\odot \vct{w_d}(i,j)}}\label{eq:weight} \\
	& \tensor{Z}=\textit{F}_{\substack{1\leq m\leq M\\1\leq n\leq N\\1\leq d\leq N_d}}(\{ \tensor{U_d}(m,n)\})\label{eq:fusion}\\
	& \nonumber d \in[1,\dots,N_d]
\end{align}

\begin{table}[t]
	\caption{Atrous Vs Normal Convolution Savings}
	
	\begin{center}
		
		\begin{tabular}{l|c|c} 
			\textbf{Type} & \textbf{Effective Receptive Field} & \textbf{Parameters}\\
			\hline
			Normal Depthwise & $3\times3$ & $9$\\
			\hline
			Atrous Depthwise & $3\times3$ & $9$\\
			
			\hline
			Normal Depthwise & $5\times5$ & $25$\\
			\hline
			Atrous Depthwise & $5\times5$ & $9$\\
			
			\hline
			Normal Depthwise & $7\times7$ & $49$\\
			\hline
			Atrous Depthwise & $7\times7$ & $9$\\
			
			\hline
			Normal Depthwise & $11\times11$ & $121$\\
			\hline
			Atrous Depthwise & $11\times11$ & $9$\\
		\end{tabular}
	\end{center}
	\label{tab:dilatedconv}
\end{table}

\subsection{Macro-Architecture Design Choices}\label{macto-architecture}
The ACFF macro block is used as a starting point to build a deep neural network that is characterized by low-computational complexity and is suitable for embedded platforms. The following design choices are made for the overall network structure which is illustrated in Table \ref{tab:EmergencyNet}.

\begin{itemize}
	\item \textbf{Reduced Cost of First Layer}: The first layer typically incurs the higher computational cost since it is applied across the whole image. Hence, a relatively small number of filters is selected ($16$) with spatial resolution of $3\times 3$. A standard convolution layer is used here to better capture low-level image features. A stride of $2$ is used to reduce the computations.
	\item \textbf{Early downsampling}: Downsampling is performed at all the initial layers. A combination of stride and max-pooling layers are used in an effort to reduce the loss of information by aggressive striding, but still reduce the spatial resolution. It was empirically found that downsizing the feature maps in the latter stages resulted in decreased accuracy hence, the downsampling is performed in the first four layers.
	\item \textbf{Canonical Architecture:} To keep the representational expressiveness a pyramid-shaped form is adopted for the CNN configuration, which means a progressive reduction of spatial resolution of the feature maps at each layer with an increase of their depth. It is quite typical for large networks to have even thousands of filters at each layer, however, for embedded applications this adds considerable overhead. Hence, the first layer has $16$ filters, which are then increased thereafter but does not go beyond $256$ which is the final layer prior to the classification part. 
	\item \textbf{Fully Convolutional Architecture:} A simple and effective trick is utilized to massively reduce the parameter count and computational cost by avoiding the use of fully connected layers. Instead, a channel-wise $1\times1$ convolution is used to reduce the channels to the number of classes, followed by a global average pooling operation that summarized the per class feature maps and which is fed to a softmax layer.
	\item \textbf{Capped leaky relu:} A capped version of ReLU is used, similar to \cite{MobileNetsV2.2018}, as the main non-linearity across the network due to its robustness when used with low-precision computation. Specifically, the output is upper bounded at $255$ which can be approximated with an $8$ bit integer. Even though quantization is not not exploited here, by utilizing this ReLU variant, the network is amicable to further improvements. Moreover, the modified capped leaky ReLU has two modes of operation. During training phase it allows the small fraction of gradient to flow, whereas in the inference phase that part is zeroed out allowing for quantization if necessary.
	\item \textbf{Regularization:} Due to the relatively small size of the dataset compared to databases such as ImageNet; additional regularization techniques are also incorporated beyond augmentation to combat overfitting. In particular, \textit{batch normalization} \cite{BatchNorm:Ioffe:2015} is used after the atrous convolutions, $1\times1$ convolutions, as well as depth reduction operations. A \textit{dropout} layer with rate $0.2$ is applied after the layers which have the highest number of parameters. As with other works L2 weight regularization with a parameter of $5\times10^{-4}$ was found to be effective in helping the network learn faster and achieve lower error rates.ca
	\item \textbf{Network Depth:} Deep networks are necessary to build strong representations but are also predicated on having a huge amount of data. Also very deep networks incur a higher computational cost. Given these two factors it was empirically found that a network size of $6$ ACFF blocks was sufficient to achieve comparable accuracy to the sate-of-the-art, while increasing it did not result in significant accuracy improvements but incurred higher computational cost.
	\item \textbf{Skip Connection Fusion:} A significant amount of information can be lost during the aggressive downwsampling process. To preserve some of the initial information the input feature map is also included into the fusion process.
\end{itemize}

The aforementioned configurations are combined to build the base \textit{EmergencyNet} architecture shown in Table \ref{tab:EmergencyNet}. Different modes of fusing together the feature maps have been evaluated along with base implementations of standard convolution, depth-wise convolution networks, and spatially-separable convolutions in order to compare and contrast the trade-offs. These results are presented in section \ref{subsec:analysis}.

\begin{table}[t]
	\caption{EmergencyNet Model Structure}
	
	\begin{center}
		
		\begin{tabular}{l|c|c|c|c} 
			\textbf{Layer} & \textbf{Output} & \textbf{Receptive} & \textbf{Number of} & \textbf{Stride}\\
			& \textbf{Size} & \textbf{Field} & \textbf{Filters} &\\
			\hline
			\textit{Input image} & $240\times240$ &  &  & \\
			\hline
			\textit{Convolution} & $120\times120$ & $3$ & 16 & $2$\\
			\hline
			\textit{ACFF Block 1} & $120\times120$ & $3,5,7$ & $64$ & $1$\\
			\hline
			\textit{MaxPooling 1} & $60\times60$ & $2$ & & $2$\\
			\hline
			\textit{ACFF Block 2} & $60\times60$ & $3,5,7$ & $96$ & $1$\\
			\hline
			\textit{MaxPooling 2} & $30\times30$ & $2$ & & $2$\\
			\hline 
			\textit{ACFF Block 3} & $30\times30$ & $3,5,7$ & $128$ & $1$\\
			\hline
			\textit{MaxPooling 3} & $15\times15$ & $2$ & & $2$\\
			\hline
			\textit{ACFF Block 4} & $15\times15$ & $3,5,7$ & $128$ & $1$\\
			\hline
			\textit{ACFF Block 5} & $15\times15$ & $3,5,7$ & $128$ & $1$\\
			\hline
			\textit{ACFF Block 6} & $15\times15$ & $3,5,7$ & $256$ & $1$\\
			\hline
			\textit{Convolution} & $15\times15$ & $1$ & $5$ & $1$\\
			\hline
			\textit{Global Pooling} & $5$ &  & & \\
			\hline
			\textit{Softmax} & $5$ &  & $5$ & \\
		\end{tabular}
	\end{center}
	\label{tab:EmergencyNet}
\end{table}
	
\begin{algorithm}[t]
	\caption{Algorithm pseudocode for balanced training}
	\label{alg:train}
	\begin{algorithmic}[1]
		\STATE  \textbf{Inputs: }$batch\_size =64, num\_of\_clastses = 5, cnn\_model$\\
		\STATE \textbf{Output: }Trained CNN Model
		\STATE \textbf{function} \textit{Train}($cnn\_model,batch\_size,num\_of\_classes$)\:
		\STATE $amount = \lceil \frac{batch\_size}{num\_of\_classes}\rceil$\\
		\COMMENT{ \% When the $batch\_size$ is not divisible to the $num\_of\_classes$ some class will be chosen each round to have an additional sample at random.} \\
		\WHILE{\textit{Training}}
		\STATE $Batch = \left[ \right]$
		\FOR{class $i$} 
		\STATE $class\_images \leftarrow$ get\_class\_samples($i$,$amount$)
		\STATE $class\_images \leftarrow$ augment\_images($class\_images$) 
		\STATE $Batch \leftarrow$ add\_to\_batch($class\_images$) 
		\ENDFOR
		\STATE cnn\_model$\leftarrow$ fit(\textit{cnn\_model},$Batch$) 
		\ENDWHILE
		\STATE \textbf{return} \textit{cnn\_model}

	\end{algorithmic}
\end{algorithm}

\subsection{Training}\label{subsec:training}
All the networks are developed and tested through the same training framework so as to have the same conditions and a fair comparison during the inference phase. The Keras deep learning framework \cite{keras} is used which has available all models with Tensorflow \cite{tensorflow} running as the backend. The same image size is used for all networks where possible (except for the \textit{mobileNet} V1 and V2 models which specifically require a smaller image size). Consequently, before augmenting and adding an image to the batch it is first resized to the appropriate image size depending on the network (default is $240\times240$ pixels which is a typical size for training CNNs). It should be noted that it is possible to use larger image sizes at a cost of slower inference time,  however in this work the image size space is not explored but rather focus is on the network design.

The first step in the training process is to split the dataset into training, validation, and test sets. The bulk of the data are allocated to the training set and the rest between the other two sets in a $0.5, 0.2, 0.3$ ratio. As mentioned prior, the \textit{Normal} class is the majority class and thus is over-represented in the dataset. This reflects real-world conditions, however, if not addressed, it can potentially lead to problems where the network overfits and thus classifies everything as the majority class. To avoid issues due to the dataset imbalance the simultaneous use of majority class undersampling with oversampling of the minority classes within the same batch is performed. To do this the process outlined in Algorithm \ref{alg:train} is performed to select the same number of images form each class to form a batch and this way all cases are equally represented. Another reason why this process is preferred is that the majority of images were extracted from videos. Even if the videos are sparsily sampled there exists a dependency between images in close temporal proximity. Hence, through random sampling it is less frequent to encounter visually similar images during the training process.

All the networks where trained using a GeForce Titan Xp, on a PC with an Intel $i7-7700K$ processor, and $32$GB of RAM. The standard Adam optimization method \cite{Adam:2014} was used for training with a cosine learning rate annealing schedule. The initial learning rate was set to $0.1$ and the total amount of epochs is $300$, thus the learning rate $LR$ at time \textit{t} is calculated as shown in Eq. \ref{eq:cosLR}. During training label smoothing is also applied with an \textit{epsilon} of $0.1$. Each epoch runs for $60$ iterations with a batch size of $64$ resulting in a total of $3,840$ augmented images per epoch.

\begin{align}
	LR(t) = 0.5 * (1 + \cos(((t * \pi) / (300)))) * 0.1
	\label{eq:cosLR}
\end{align}

\section{Experimental Evaluation and Results}\label{sec:results}
In  this  section, the experimental evaluation of the proposed methodology is discussed with results from the experimental evaluation of the approach on an actual embedded platform. First the improvements over existing networks are validated on the developed dataset and qualitative results of the learning process are presented. Real experiments  have also  been conducted  in  two  different  settings: (i)  On-board  embedded  processing where all  the  computations  are  performed  on-board  the  resource-constrained  UAV  platform,  on  an  embedded  device. (ii) Remote-based processing in  which  the  UAV  transmits  the  captured video  to the controller ground station for  processing on an Android tablet that controls the UAV.  It is worth noting that the primary interested is in single image processing speed and as such the evaluation phase is carried our with a batch size is $1$, since this is common in real-time streaming applications where the camera outputs each frame sequentially.

\subsection{Performance Metrics}\label{subsec:metrics}
An important performance metric for real-time applications is the resulting frame-rate or frames-per-second (FPS) achieved by each model, which is inversely proportional to the time needed to process a single image frame from a video/camera stream. In addition, since the prior distribution over classes is signiﬁcantly nonuniform a simple accuracy measure (percentage of correctly classiﬁed examples) which is used in related works, may not be appropriate for the specific problem considered in this work since usually the normal case is much more frequent than other classes. To avoid this bias in our results the F1 score \cite{Evaluation:Metrics:Classification:2015} is employed as the learning performance metric instead. The key metrics are summarized below:

\subsubsection{Frames-Per-Second (FPS)}\label{subsubsec:fps}
The rate at which a classifier is capable of processing incoming camera frames, where $t_i$ is the processing time of a single image. 

\begin{align}
	FPS_{model} = \dfrac{1}{\dfrac{1}{N_{test\_samples}}\times\sum_{i=1}^{N_{test\_samples}}(t_i)},  \label{eq:fps}
\end{align}

\subsubsection{Mean F$1$ Score ($\overline{F1}$)}\label{subsubsec:map}
The F1 score metric finds the mean accuracy across all classes weighted by the number of test instances for each label denoted as $|C_i|$. F1 score conveys the balance between the precision and the recall. In general a good F1 score means that you have low false positives and low false negatives. This handles the label imbalance problem and provides a more appropriate measure of the performance across classes.

\begin{align}
	\overline{F1}_{model} = \dfrac{2}{N_{classes}}\times\sum_{i=1}^{N_{classes}} \dfrac{prec_i\times sens_i}{prec_i+sens_i},  \label{eq:f1}  
\end{align}

\subsection{Overall Performance Analysis and Comparison}\label{subsec:analysis}
Initially, the custom models with different fusion mechanisms with the main architecture presented in Section \ref{subsubsec:custom} are compared against vanilla CNN implementations of standard convolution networks as well as networks composed of depthwise- and spatially- separable convolutions. The different networks are compared against parameter size which determines the learning complexity of the model, memory for storage of the model weights, and accuracy with the results summarized in Table \ref{tab:models}.

First the standard convolutional neural network has the higher amount of parameters and subsequently memory demands compared to other models. It achieved the highest accuracy amongst the standard networks with the highest demands for memory and parameter count. The depthwise-separable convolution network reduces these demands however, it has a considerable accuracy drop. The spatially-separable network increases the accuracy but with additional parameters needed. On the other hand, the proposed networks with different fusion mechanisms provide higher or on par accuracy. due to the limited number of parameters they are computationally and memory efficient while the combination of multiple features from different kernel sizes manages to provide a considerable accuracy increase. The \textit{maximum} and \textit{add} fusion networks perform better for this particular dataset, amongst the ACFF-based networks. This can be attributed to the fact that they give more emphasis to specific spatial locations which helps in the learning rather than \textit{average} fusion which suppresses information and may reduce some important features. Also they do not increase the memory and processing requirements which is the case in \textit{concatenation} fusion. The best performing network with \textit{add} fusion is selected as the main architecture of \textit{EmergencyNet} which is used for comparison with other networks and further experimental results.

The results for all networks are summarized in Table \ref{table:model_res}. For these networks comparisons are also made with respect to Floating Point Operations (FLOPS) to measure complexity as a platform independent metric. First, with regards to the accuracy of the pretrained models it is observed that \textit{VGG16} outperforms all of them with a $96.4\%$ F1 accuracy score with \textit{ResNet50} closely following with $96.1\%$. This is in line with what has been reported in prior works using such networks achieving accuracies between $81-98\%$ for different applications and scenarios. However, both networks have very high demands for computational and storage requirements making them unsuitable for resource constraint systems and real-time use. The latest iteration V3\footnote{The smaller varient of MobileNet V3 was used as it is more computationally efficient.} achieves the higher accuracy of the mobilenet family of networks it has the lowest FLOPs from the pretrained networks and achieves a score of $95.3\%$. It requires an order of magnitude more parameters and memory, however. The other mobilenet versions (V1 and V2) operate on smaller image resolutions ($224\times224$), however still have higher FLOP requirements. \textit{EfficientNet}\footnote{The optimized version B0 is used} provides a high accuracy of $96.0\%$ due to its elaborate architecture. However, the parameter count and memory requirements are higher than EmergencyNet. Other networks fail to provide adequate accuracy and require a higher number of FLOPs with more parameters. It is clear from this analysis that it is worth investigating tailored made solutions for constrained applications in order to provide an improvement across all design aspects.
	
\textit{EmergencyNet} achieves the same accuracy with fewer FLOPs and parameters than the most competitive architectures. It manages to achieve $95.7\%$ F1 score accuracy which is very close to the pretrained network approaches with minimal memory requirements at $\sim360KB$ which is $\sim30\times$ smaller than \textit{MobileNetV3}. Overall, \textit{EmergencyNet} has comparable performance to more recent models such as \textit{EfficientNet} and \textit{MobileNetV3} in terms of accuracy with much less parameters. In addition, the significantly smaller model size of \textit{EmergencyNet} compared to other models illustrates its efficacy for greatly reducing the memory requirements for leveraging image classification for real-time embedded emergency response applications, which also makes it suitable for on-chip storage on low-power platforms with limited memory as well as more specialized computing platforms such as FPGAs which can have limited on-chip storage \cite{Kyrkou:2016:TNNLS}.

\begin{table}[t]
	\caption{Comparison of the different fusion mechanisms and base models}
	
	\begin{center}
		
		\begin{tabular}{l|c|c|c} 
			\textbf{Model} & \textbf{Parameters} & \textbf{Memory} & \textbf{F1 Score}\\
			&  & \textbf{($MB$)}  & \textbf{(\%)} \\
			\hline
			Standard Convolutional & $719,737$ & $3.08$ & $93.1$ \\
			\hline
			Depthwise-Separable & $95,849$ & $0.383$ & $90.5$ \\
			\hline
			Spatially-Separable & $627,913$ & $2.511$ & $92.5$ \\
			\hline
			Max-Fusion & $90,892$ & $0.363$ & $95$ \\
			\hline
			Average-Fusion & $90,892$ & $0.363$ & $93.9$ \\
			\hline
			Add-Fusion  & $90,892$ & $0.363$ & $95.7$ \\
			\hline
			Concatenate Fusion & $222,881$ & $0.891$ & $94.3$ \\
		\end{tabular}
	\end{center}
	\label{tab:models}
\end{table}

\begin{table}[t]\centering
	\begin{threeparttable}
		
		\caption{Comparison with existing approaches}
		\label{table:model_res}
		
		\begin{tabular}{c|c|c|c|c}
			\textbf{Model} & \textbf{Parameters} & \textbf{Memory} &\textbf{F1 Score}  & \textbf{FLOPS }\\ 
			&  &  \textbf{(MB)} & \textbf{(\%)} &  ($\times10^6$) \\
			\hline
			\textit{EmergencyNet}
			& 90,892 & 0.368 & 95.7 & 57 \\ \hline
			\textit{ResNet50}
			& 24,113,541 & 96.4 & 96.1 & 4,533 \\ \hline
			\textit{VGG16}
			& 14,849,349 & 59.39 & 96.4 & 17,620 \\ \hline
			\textit{Xception}
			& 21,387,309 & 85.549 & 95.3 & 419 \\ \hline
			\textit{MobileNet V1}
			& 3,492,549 & 13.9 & 95 & 550 \\ \hline
			\textit{MobileNet V2}
			& 2,587,205 & 10.3 & 95.2 & 279 \\ \hline
			\textit{MobileNet V3}
			& 3,046,037 & 12.1 & 95.3 & 60 \\ \hline
			\textit{SqueezeNet}
			& 698,917 & 2.7 & 91.5 & 833 \\ \hline
			\textit{ShuffleNet}
			& 4,282,425 & 17.1 & 91.1 & 524 \\ \hline
			\textit{EfficientNet (B0)}
			& 4,378,785 & 17.5 & 96.0 & 420 \\ \hline
			\textit{\textit{Fire\_Net}}
			& 5,235,860 & 5.2 & 90.5 & 1577\\
			
		\end{tabular}
	\end{threeparttable}
\end{table}

\begin{figure}[H]
	\centering
	\includegraphics[width=0.5\columnwidth]{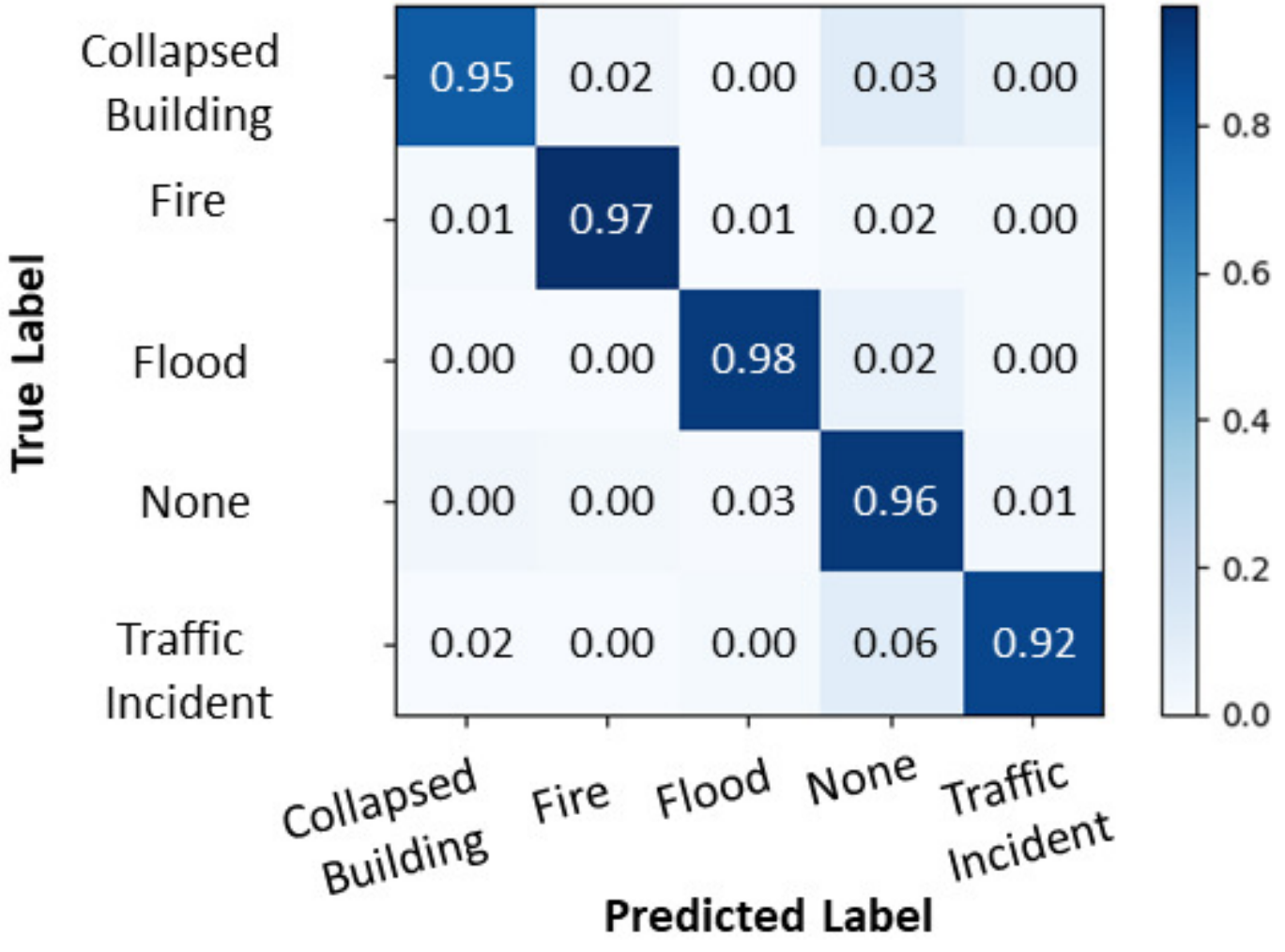}
	\caption{EmergencyNet Confusion Matrix}
	\label{fig:cnf_matrix}
\end{figure}

\begin{figure}[H]
	\centering
	\includegraphics[width=0.99\columnwidth]{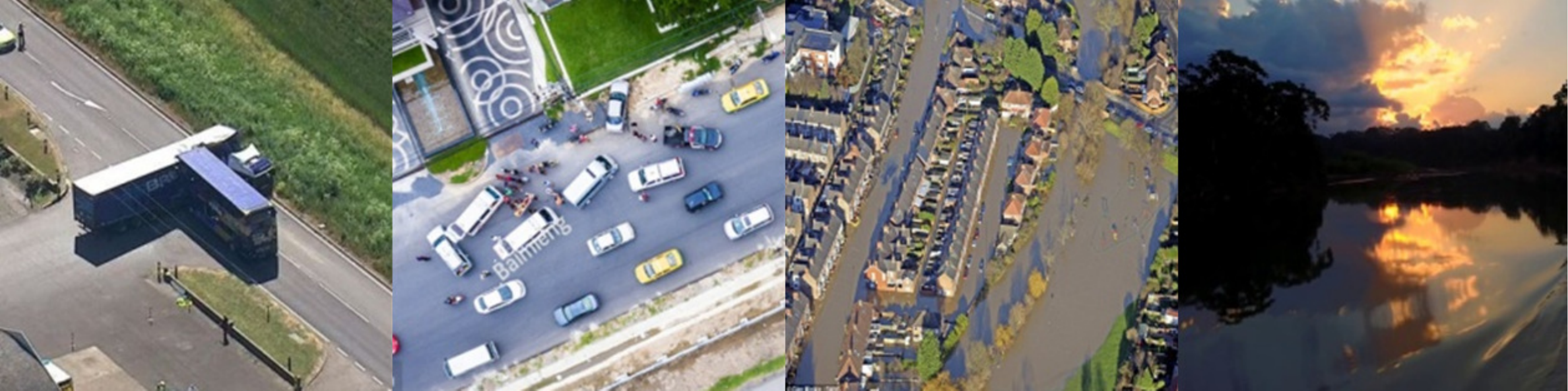}
	(a)\hspace{50pt}(b)\hspace{50pt}(c)\hspace{50pt}(d)
	\caption{Interesting Classification Cases:(a \& b) In these cases the incident let that be crash or damage is not clearly visible and as a results the models mis-classify the images. (c) Image of flood not classified correctly as the hue is similar to road. (d) The orange-yellow hue does not confuse the model as it correctly classifies the image}
	\label{fig:cases}
\end{figure}

The confusion matrix in Fig. \ref{fig:cnf_matrix} of \textit{EmergencyNet} allows us to draw some conclusions regarding the difficulty of the task and also highlight some potential research opportunities. Some areas of improvement can come from reducing the majority of errors attributed to classifying images as being in the \textit{normal} class. This is to be expected since there are some images where the incident does not occupy a large image area and the viewpoint may not be ideal. To atone for such cases it is a common practice to process multiple crops from the input images. However, this has negative implications with regards to the processing time. Also for the \textit{traffic incident} class it is observed that overall it produced the lowest accuracy. This is because some of the images the "incident" was not clear either because of the viewpoint or because in some cases the label was ambiguous. Some examples of interesting classification outcomes are shown in Fig. \ref{fig:cases}. It is worth noting that these behaviours were consistent across all networks, agreeing with recent studies that in difficult circumstances more contextual information is needed in order for CNNs to make better predictions.

\subsection{Qualitative Evaluation of Learning}\label{subsec:learn_eval}
\begin{figure}[t]
	\centering
	\includegraphics[width=0.7\columnwidth]{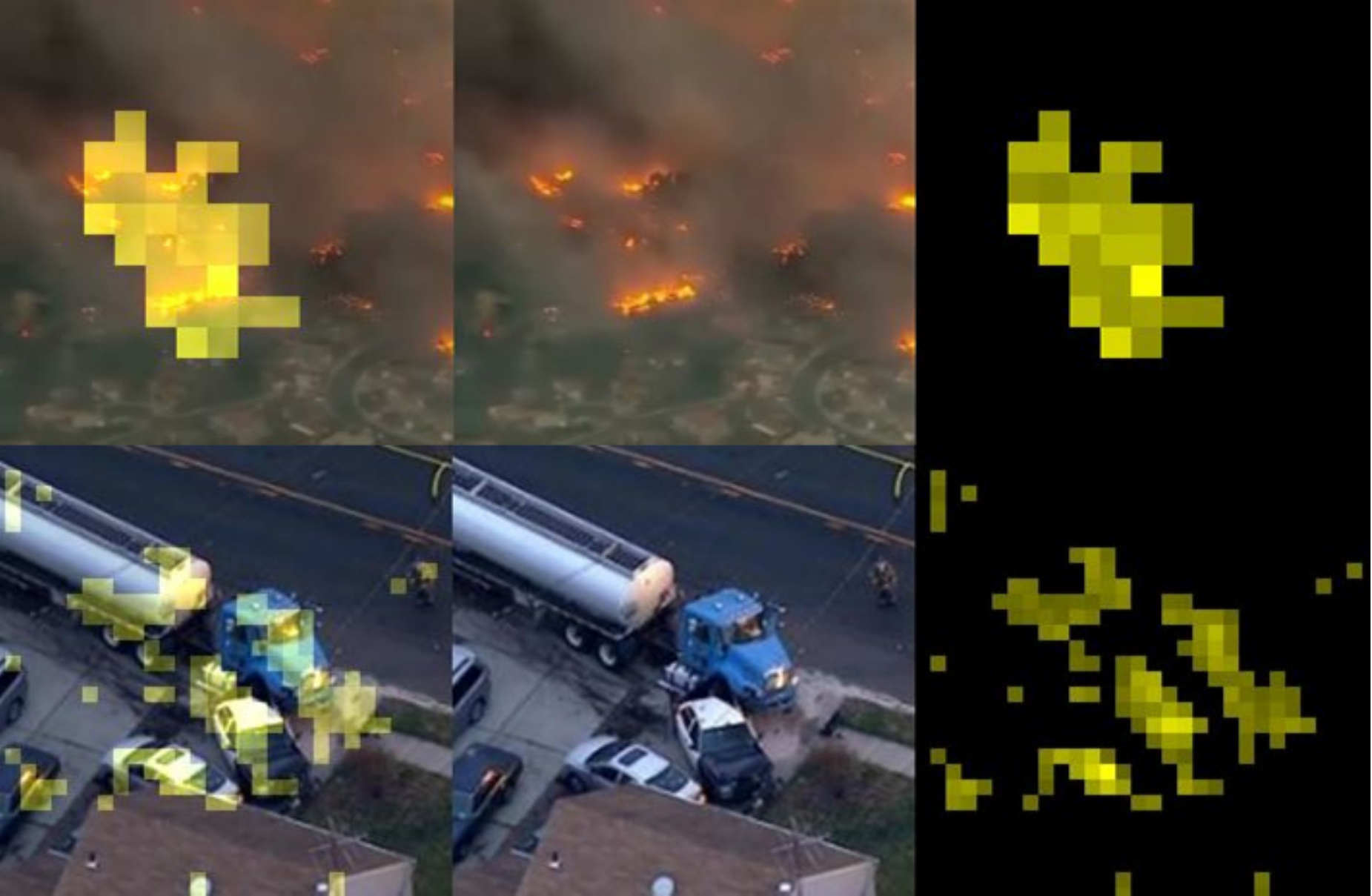}
	\caption{(Left) Feature Maps overlaid on input images (center) input images (right) High Activation feature Maps: Image regions that correspond to the highest activations in the network. These areas are mainly concentrated on the main characteristics of each event. In all cases the network focuses on important cues within the image to makes its prediction. For example in the first image it focuses on the red-orange glow of fire as well as the rubble of the collapsed building in the third image. }
	\label{fig:nvidia_viz}
\end{figure}

\begin{figure}[t]
	\centering
	\includegraphics[width=0.6\columnwidth]{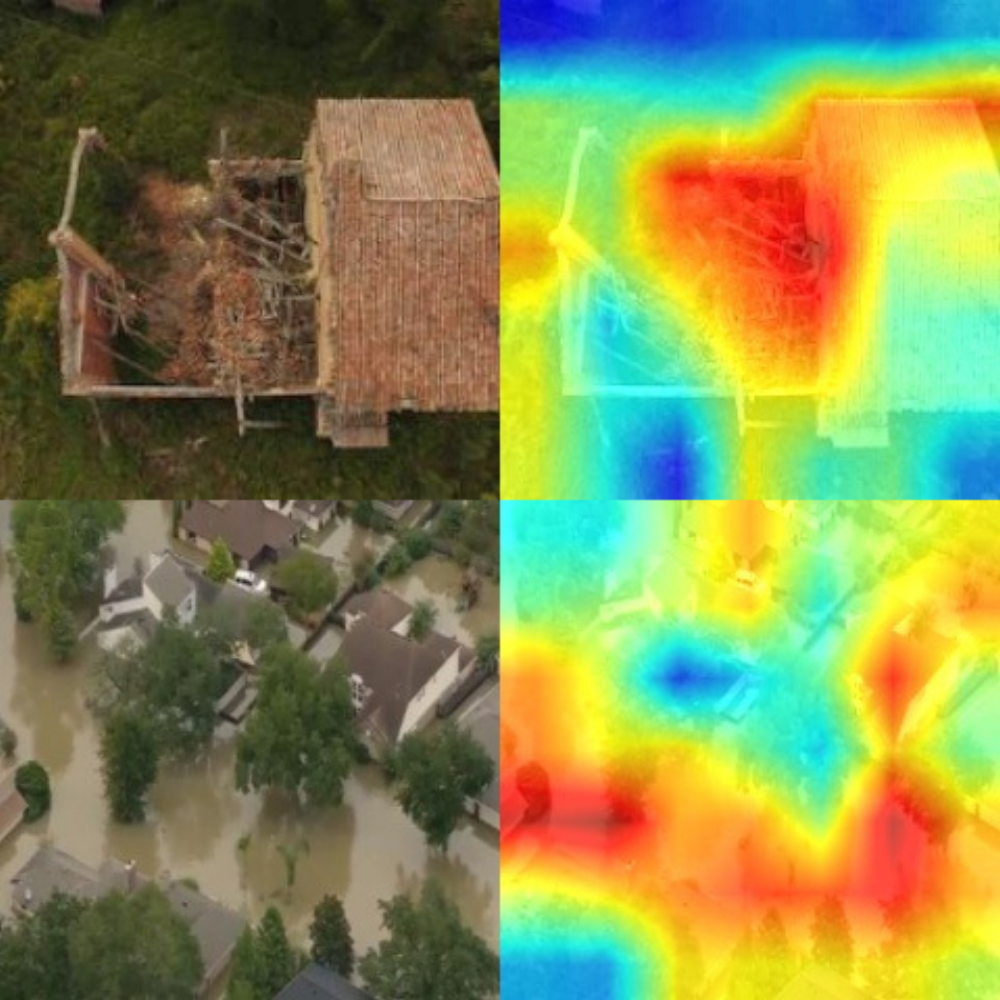}
	\caption{Images classified correctly and the corresponding class activation map. In all cases the visualization shows that the network focuses on important cues within the image to make a decision. (top) Focuses primarily on the rubble of the collapsed building. (bottom) Focuses on the flooded area.}
	\label{fig:gradcam_viz}
\end{figure}

In this section a closer look is taken into what features and image regions influence the prediction of the neural network and what it has learned to respond to in the various cases in order to come to a classification decision. To this end, the approach in \cite{NVIDIA:ExplainingDNNCar:Bojarski:2017} is used to find parts of the image that produce the highest activations. Through this approach in each layer, the activations of the feature maps are averaged and are scaled up to the size of the map of the preceding layer. The  up-scaled averaged map from an upper level is then multiplied with the averaged map from the layer below. These steps are repeated until the input is reached. In addition, the Gradient-weighted Class Activation Mapping (Grad-CAM) approach \cite{Selvaraju.CAM.2017} is also employed to visualizing the regions of input that are important for class-specific predictions and uses gradient information flowing into the final convolutional layer rather than activation output. These visualizations mask shows which regions of the input image contribute most to the output of the network. Figure \ref{fig:nvidia_viz} indicates some images which have been correctly classified and the regions corresponding to the highest activation's of the CNN (the \textit{EmergencyNet} model in particular). Fig. \ref{fig:gradcam_viz} indicates some images which have been correctly classified and the produced heat-map indicating important regions in the CNN's decision making. Through both approaches it is evident that the CNN uses important cues in order to come to a decision such the red-yellow glow fire in Fig. \ref{fig:nvidia_viz} and the building rubble in Fig. \ref{fig:gradcam_viz}. The heat-maps consistently show that the pixels corresponding to the disaster event dominate the importance of the classification result. This indicates that the network indeed learns how to spot important incidents for emergency situations. 

\begin{figure*}[t]
	\centering
	\includegraphics[width=1\linewidth]{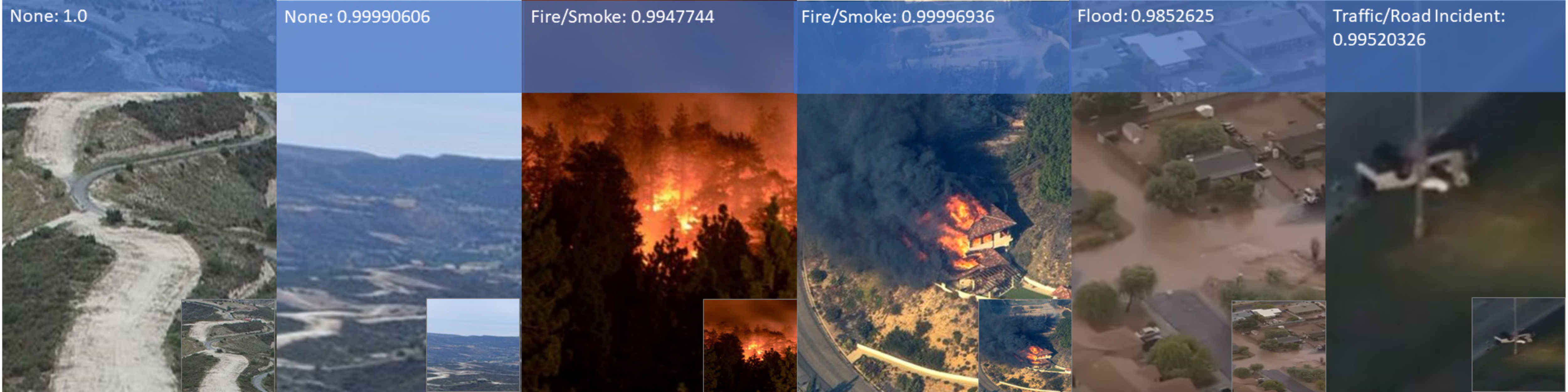}
	\caption{Experimental embedded platforms: Screenshots from Android mobile ground station. Neural network confidence and prediction outcome at the top of each image.}
	\label{fig:embedded_platforms_android}
\end{figure*}

\begin{figure*}[t]
	\centering
	\includegraphics[width=0.99\linewidth]{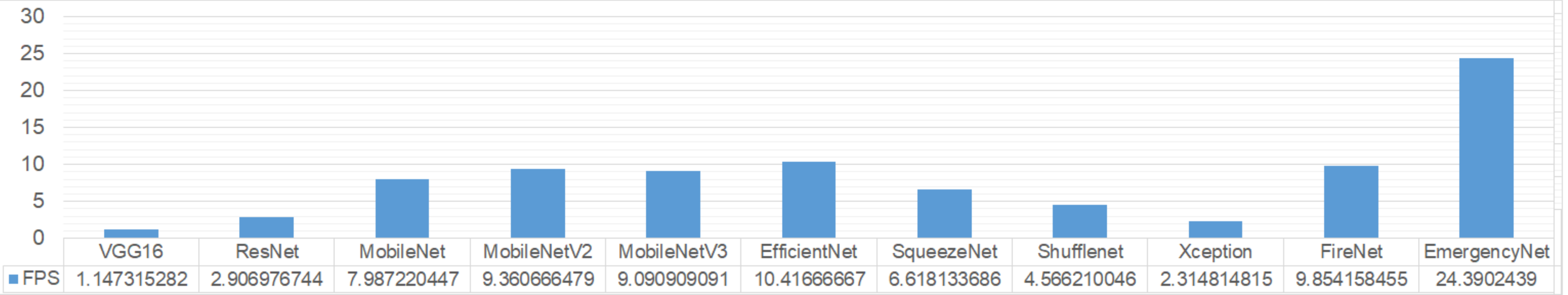}
	\caption{Frames-per-second when running on the ARM-based platform.}
	\label{fig:embedded_fps}
\end{figure*}

\subsection{Embedded Application Results}\label{subsec:embedded}

\begin{figure}[t]
	\centering
	\includegraphics[width=0.6\linewidth]{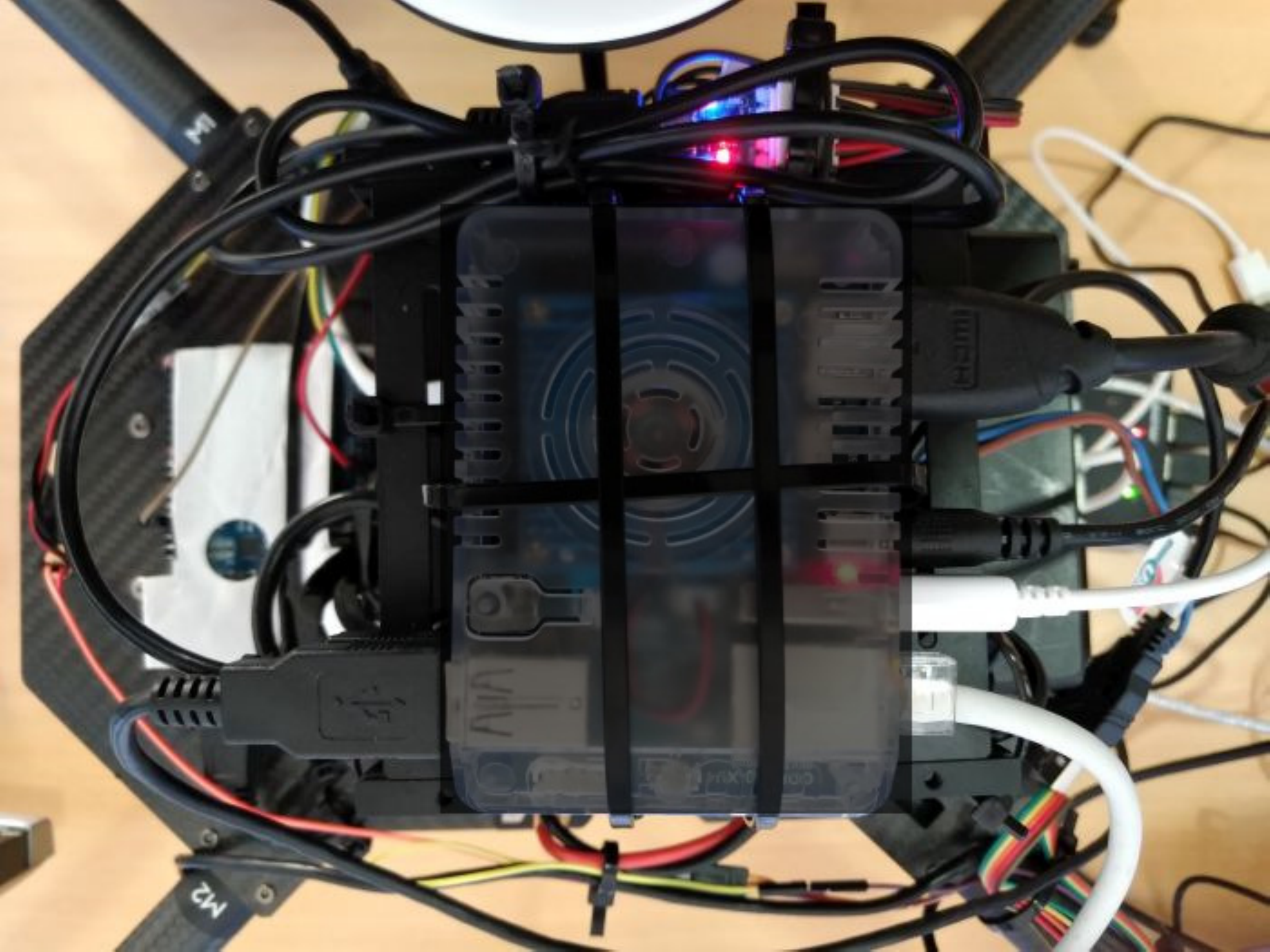}
	\caption{Experimental embedded platforms on-board a DJI Matrice 100 UAV}
	\label{fig:embedded_platforms_odroid}
\end{figure}

This section presents the evaluation of \textit{EmergencyNet} in real use-cases. Edge devices with limited computational resources and restrictive energy overhead are targetted such as ARM-based mobile devices with 10-150 MFLOPs. The processing speed and subsequent frame-rate are measured for the proposed network, \textit{EmergencyNet}, along with other state-of-the-art networks. Two different scenarios are targetted. The first concerns an on-board processing platform and the second a tablet which acts as the UAV mobile ground station that is connected to the UAV controller to process the input image. Both options are easily deployable and thus suitable for remote monitoring in emergency scenarios. Experimental setup using a DJI Matrice $100$ UAV and experimental results for the two use-cases are shown in Fig. \ref{fig:embedded_platforms_odroid} and Fig. \ref{fig:embedded_platforms_android} respectively. 

\subsubsection{UAV On-board Processing}\label{subsubsec:odroid}
For the on-board processing a platform featuring with a quad-core ARM Cortex-A57 (Fig. \ref{fig:embedded_platforms_odroid}) is used which is powerful and energy-efficient and comes in a small form factor suitable for UAVs. As shown in Fig. \ref{fig:embedded_fps} the \textit{EmergencyNet} model achieves $\sim25$ FPS on this platform which is already far more practically applicable than other state of the art models that achieve at most $\sim9$ FPS. It is important to note that the reduced FLOPs obtained by \textit{MobileNetV3} does not directly translate to improved performance on the specific CPU-based experimental platform. This further emphasizes the need to optimize with the specific characteristics of the processing platform in mind. Overall, a speedup of $2-20\times$ is observed compared to other models. However, it is possible to provide further gains by applying quantization and bit reduction techniques, which \textit{EmergencyNet} is already well suited for, to further improve performance for CPU platforms and potentially run at even higher frame-rates.

\subsubsection{Processing on UAV Mobile Ground Station}\label{subsubsec:android}
Evaluation is also performed on a more conventional system where the UAV controller is connected to a tablet which acts as a mobile control station that is fed with the UAV camera image (Fig. \ref{fig:embedded_platforms_android}). In this setup the video feed can be directly processed on the tablet without impacting the UAV platform in terms of weight and power. The \textit{EmergencyNet} network manages to offer higher frame-rates $\sim19$ FPS compared to state-of-the-art networks which remain below $10$ FPS. Notice, that in real-world cases, interference can be observed between the UAV and the controller thus negatively impacting the reliability of the visual task. Hence the former case can be a more robust option. 

\subsection{Discussion and Further Improvements}\label{subsec:embedded}
Overall,  the  results  are  encouraging  in  that  the \textit{EmergencyNet} network  is able  to provide accuracy comparable to state-of-the-art models at a fraction of memory and computation. There are still improvements that are possible particulary if the computing platform allows for them. The large margins afforded by \textit{EmergencyNet} allows us to process even higher resolution images. Particularly, by combining the efficient model with tiling approaches that break larger images into smaller regions that can be processed individually by a model. Especially, by combining them into a single batch and utilizing parallelization capabilities it is possible to offer comparable frame-rates for higher resolution images \cite{Plastiras:2018:ECO:3243394.3243692}.

With regards to video streams the overall classification performance can be further improved through the assumption that subsequent frames in a video are correlated with respect to their semantic contents. By measuring the distance between the output vectors prior to the classification layer it is possible to infer the similarity between different frames and accordingly weight the predictions of the past frames in order to remove any classification flickering and smooth out the predictions. In our test videos this simple yet effective technique managed to reduce prediction flickering.

The feature fusion architecture was found to provide good representational power with reduced parameter count. Hence, it would be beneficial to exploit recent advancements in automated model search which are being increasingly deployed in recent works\cite{Howard_2019_ICCV,EfficientNet:2019} as a logical next step to further improve the feature fusion architecture.

Through the analysis some difficult cases have been identified that require further research in order to develop better solutions. For example, the \textit{traffic incident} case can cover a broad range of scenarios and breaking down into subclasses can be beneficial. Another important challenge was the difficulty of collecting and gathering the data. For particular cases where it is not easy to gather loads of data through the internet it could be beneficial to explore the recent advances in generative models to learn the joint probability distribution for each class in order to generate novel realistic synthetic data. 

Finally, the potential to combine \textit{EmergencyNet} with algorithms that detect people and vehicles as well as additional modalities (e.g., infrared or thermal cameras) can lead to even more enhanced situational awareness that can provide valuable tool for emergency response and disaster management applications especially when integrated with geospatial applications for geo-tagging of the recognized events.

\section{Concluding Remarks}\label{sec:conc}
This paper has been a foundation study of on-board UAV processing for emergency response applications. An analysis on the design and implementation of an efficient deep learning system has been carried out to automatically recognize and classify disaster events in real-time from on-board a UAV. 
The proposed solution provides an adequate trade-off between accuracy, inference speed, and complexity and that it can be used as a building block towards use-cases with similar constraints.
The experimental study validates the efficiency of the proposed method since \textit{EmergencyNet} is up to $20\times$ faster, requires an order of magnitude less memory and provides similar or better accuracy to existing models. In addition, a dedicated aerial image dataset for emergency response applications is introduced which researchers can use to further advance the existing models. The dataset will be further expanded and enhanced with additional images and classes in order to further raise the awareness of the community towards such applications and improve on existing models and techniques.

\section*{Acknowledgment}
This work is funded by European Union's Horizon 2020 research and innovation programme under grant agreement No 739551 (KIOS CoE) and from the Republic of Cyprus through the Directorate General for European Programmes, Coordination and Development.

Christos Kyrkou gratefully acknowledges the support of NVIDIA Corporation with the donation of the Titan Xp GPU used for this research. 

\bibliographystyle{plain}
\bibliography{./citations}

\end{document}